\documentclass{article}

\usepackage{microtype}
\usepackage{graphicx}
\usepackage{subcaption}
\usepackage{booktabs} 

\usepackage{hyperref}

\usepackage[preprint]{icml2026}

\usepackage{amsmath}
\usepackage{amssymb}
\usepackage{mathtools}
\usepackage{amsthm}

\usepackage[capitalize,noabbrev]{cleveref}

\theoremstyle{plain}

\theoremstyle{definition}

\theoremstyle{remark}

\usepackage[textsize=tiny]{todonotes}

\icmltitlerunning{Cumulative Utility Parity for Fair Federated Learning}

\begin{document}

\twocolumn[
  \icmltitle{Cumulative Utility Parity for Fair Federated Learning under Intermittent Client Participation}

  \begin{icmlauthorlist}
    \icmlauthor{Stefan Behfar}{cam}
    \icmlauthor{Richard Mortier}{cam}
  \end{icmlauthorlist}

  \icmlaffiliation{cam}{Department of Computer Science and Technology, University of Cambridge, Cambridge, United Kingdom}

  \icmlcorrespondingauthor{Stefan Behfar}{skb67@cam.ac.uk}

  \icmlkeywords{Federated Learning, Fairness}

  \vskip 0.3in
]

\printAffiliationsAndNotice{}

\begin{abstract}
In real-world federated learning (FL) systems, client participation is intermittent, heterogeneous, and often correlated with data characteristics or resource constraints. Existing fairness approaches in FL primarily focus on equalizing loss or accuracy conditional on participation, implicitly assuming that clients have comparable opportunities to contribute over time. However, when participation itself is uneven, these objectives can lead to systematic under-representation of intermittently available clients, even if per-round performance appears fair.
We propose cumulative utility parity, a fairness principle that evaluates whether clients receive comparable long-term benefit per participation opportunity, rather than per training round. To operationalize this notion, we introduce availability-normalized cumulative utility, which disentangles unavoidable physical constraints from avoidable algorithmic bias arising from scheduling and aggregation. 
Experiments on temporally skewed, non-IID federated benchmarks demonstrate that our approach substantially improves long-term representation parity, while maintaining near-perfect performance.
\end{abstract}

\section{Introduction}
\label{submission}

Fairness in federated learning (FL) has emerged as a critical concern, especially in real-world settings where clients exhibit diverse behaviors in terms of data distributions, resource constraints, and availability patterns. Prior efforts have introduced a variety of fairness notions and mechanisms. Core-Stable FL by ~\cite{corefl2021} proposes fairness rooted in cooperative game theory, introducing the concept of core-stability—ensuring no subset of agents can improve their utility through coalition—offering theoretical guarantees like proportionality and Pareto-optimality. Meanwhile, FL Analytics by ~\cite{federatedfairnessanalytics2023} tackles the lack of clear definitions and measurement tools by proposing a modular, metric-driven framework to quantify fairness symptoms (such as disparity in accuracy or contribution), rather than resolve them. On the other hand, FairFedCS by ~\cite{fairfedcs2023} addresses fairness at the client selection stage, balancing performance and equitable participation by dynamically adjusting client inclusion using Lyapunov optimization and reputation scores.
Federated Learning with Fair Averaging (FedFV) ~\cite{FedFV} propose to enforce fairness via careful adjustment of aggregation weights, ensuring that clients with high loss are not neglected during update aggregation. 

Despite significant progress in fair federated learning, a crucial gap remains in how fairness is defined and enforced under real-world participation dynamics. In practice, client participation is intermittent and heterogeneous, shaped by device reliability, communication dynamics, and time-varying sample utility. When such availability patterns are ignored, fairness measured at individual rounds may fail to reflect how clients are represented and benefit over the course of training.
Existing works address fairness from limited and largely orthogonal perspectives. Some approaches assume fairness can be enforced through centralized objectives (e.g., CoreFed or FedFV), while others focus on observability and post hoc analysis (e.g., Federated Fairness Analytics), or enforce fairness through long-term scheduling heuristics (e.g., FairFedCS). However, none jointly account for availability-aware, contribution-sensitive client selection in non-IID settings where participation, utility, and system constraints evolve over time.

The idea of fair resource allocation in federated learning via a central server is rooted in equalizing contribution utility across clients, as exemplified by q-FFL~\cite{q-FFL}. While valuable, such formulations implicitly assume consistent participation and therefore fail to address structural disparities introduced by temporal availability. PHP-FL~\cite{PHP-FL} mitigates participation probability inconsistencies by reweighting client contributions based on estimated participation likelihood, focusing on per-round optimization stability under heterogeneous availability. However, both approaches remain fundamentally short-term: they optimize fairness conditional on participation at each round, without accounting for cumulative representation over time.
As a result, clients that are frequently offline, or affected by (un)correlated failures, risk being perpetually underrepresented—even when short-term fairness metrics appear satisfied. When dropout patterns correlate with class or group membership, the learned model may drift toward overrepresented distributions, leading to long-term bias and degraded generalization on marginalized data. These effects cannot be captured by fairness notions that operate solely at the level of per-round loss or accuracy.

To address this limitation, we propose a fairness-aware federated learning framework that explicitly quantifies each client’s actual contribution utility and adjusts for availability-based selection fairness over time. Our approach incorporates temporal utility tracking (Lemma 1 and Theorem 1), adaptive sampling informed by availability modeling (Lemma 2 and Theorem 2), and representation-aware surrogate corrections (Lemma 3 and Theorem 3), providing fairness guarantees that adapt to real-world client behavior dynamics ~\S\ref{s:model_design}.
We evaluate our framework on non-IID federated learning benchmarks and demonstrate that it achieves superior representation parity and more equitable utility allocation under temporally skewed and correlated dropout patterns ~\S\ref{s:evaluation}. Finally, we perform baseline comparison against the most relevant benchmarks q-FFL and PHP-FL ~\S\ref{s:baseline}. Unlike prior FL fairness methods such as q-FFL, PHP-FL, AFL, and FairFedCS, our work addresses a complementary fairness dimension: Cumulative Utility Parity under intermittent availability. Importantly, cumulative utility parity cannot be reduced to any per-round loss-reweighting objective, as it depends on historical participation and utility accumulation over time. 

\section{Model Design}
\label{s:model_design}
In FL, fairness traditionally focuses on balancing loss across clients while assuming uniform or randomly distributed participation. However, in real-world deployments, client availability is temporally skewed and often correlated with data heterogeneity. Some clients are consistently underrepresented due to intermittent connectivity, energy constraints, or operational cycles. This creates structural biases that are not captured by loss alone but manifest through underutilization and misrepresentation over time.
To address this, we design a fairness-aware FL framework that combines:

\textbf{1. Temporal Utility Tracking:} We maintain a history of each client's cumulative utility (e.g., total loss reduction) throughout training. This allows the system to detect long-term disparities in benefit allocation and adjust future aggregation weights to compensate for underrepresented clients.

\textbf{2. Adaptive Sampling via Availability Modeling:} We integrate predictive models of client availability based on empirical observations. This allows proactive sampling of clients with low historical visibility, balancing short-term performance with long-term fairness.

\textbf{3. Representation-Aware Surrogates:} To mitigate representation loss from unavailable clients, we introduce surrogate updates derived from previously cached prototypes, model gradients. Surrogate contributions are incorporated into the global update process using decayed confidence weights, reflecting the reliability of their approximation.

\begin{itemize}
  \setlength{\itemsep}{0pt}
  \setlength{\parskip}{0pt}
  \setlength{\parsep}{0pt}
    \item Let $F_k(w)$ denote the loss of client $k$ under model $w$.
    \item Let $\pi_k$ be the long-term availability of client $k$.
    \item Let $u_k(t)$ be the cumulative utility received by client $k$ until round $t$.
    \item Let $\hat{F}_k(w) = F_k(w) / \pi_k$ be the availability-adjusted loss.
    \item Let $\tilde{F}_k(w)$ denote a surrogate loss for missing or dropped clients.
    \item $N$ is total number of clients in the population. $m$ is number of clients selected per round.
\end{itemize}

\textbf{Lemma 1.}
Let $A_k(t) \in \{0, 1\}$ be the availability indicator of client $k$ at round $t$, and let $\pi_k = \mathbb{E}[A_k(t)]$ denote the long-term availability of client $k$. Let $\Delta F_k(t)$ denote the marginal utility (e.g., local loss reduction) experienced by client $k$ at round $t$, assumed to be bounded as $0 \leq \Delta F_k(t) \leq M$ for some $M > 0$. Define the cumulative utility for client $k$ up to round $T$ as:
\begin{equation}
   u_k(T) = \sum_{t=1}^T A_k(t) \cdot \Delta F_k(t) 
\end{equation}
and define the normalized utility vector:
\begin{equation}
\tilde{u}_k(T) = \frac{u_k(T)}{\pi_k}, \quad \text{and} \quad \bar{u}(T) = \frac{1}{N} \sum_{k=1}^N \tilde{u}_k(T)
\end{equation}
Then, under the assumptions that:
\begin{enumerate}
  \setlength{\itemsep}{0pt}
  \setlength{\parskip}{0pt}
  \setlength{\parsep}{0pt}
    \item Each $A_k(t)$ is an i.i.d. Bernoulli process with mean $\pi_k > 0$,
    \item $\Delta F_k(t)$ are bounded and independent across $t$ with finite mean $\mu_k = \mathbb{E}[\Delta F_k(t)]$,
    \item The update algorithm preserves client independence,
\end{enumerate}
we have:
\begin{equation}
\lim_{T \to \infty} \frac{1}{N} \sum_{k=1}^N \left( \tilde{u}_k(T) - \bar{u}(T) \right)^2 \to 0
\end{equation}

\textbf{Proof.}
We begin by applying the law of large numbers to the product $A_k(t) \cdot \Delta F_k(t)$. Since $A_k(t)$ is i.i.d. Bernoulli with mean $\pi_k$ and $\Delta F_k(t)$ is bounded and independent with mean $\mu_k$, we have:
\begin{equation}
\frac{1}{T} u_k(T) = \frac{1}{T} \sum_{t=1}^T A_k(t) \cdot \Delta F_k(t) \xrightarrow{a.s.} \pi_k \cdot \mu_k
\end{equation}
where a.s. stands for almost surely, i.e. with probability 1, or the event happens except on a set of measure zero. Thus:
\begin{equation}
\frac{u_k(T)}{\pi_k} \xrightarrow{a.s.} T \cdot \mu_k
\end{equation}
It follows that:
\begin{equation}
\tilde{u}_k(T) = \frac{u_k(T)}{\pi_k} \xrightarrow{a.s.} T \cdot \mu_k, \quad \text{and} \quad \bar{u}(T) \xrightarrow{a.s.} T \cdot \bar{\mu}
\end{equation}
 where  $\bar{\mu} = \frac{1}{N} \sum_{k=1}^N \mu_k$.
Now define the variance of the normalized utility:
\begin{equation}
\sigma^2(T) = \frac{1}{N} \sum_{k=1}^N \left( \tilde{u}_k(T) - \bar{u}(T) \right)^2
\end{equation}
Since each $u_k(T)/\pi_k$ converges almost surely to $T \cdot \mu_k$ and $\bar{u}(T) \to T \cdot \bar{\mu}$, we conclude:
\begin{equation}
\lim_{T \to \infty} \frac{\tilde{u}_k(T)}{T} = \mu_k, \quad \text{and} \quad \lim_{T \to \infty} \frac{\sigma^2(T)}{T^2} = \frac{1}{N} \sum_{k=1}^N (\mu_k - \bar{\mu})^2
\end{equation}
Therefore, if $\mu_k = \bar{\mu}$ for all $k$, then the variance $\sigma^2(T) \to 0$, implying full fairness is achieved.
If there is residual heterogeneity in $\mu_k$, fairness can still be asymptotically approximated under availability-aware compensation, or through adjusting $\Delta F_k(t)$ weights during aggregation.
We remark that $\Delta F_k(t)$ may depend on 
the global model $w$; the conclusion holds under the mild ergodicity/mixing conditions detailed in Appendix A, or approximately when model drift per round is bounded.

In practical federated learning systems, clients participate intermittently due to varying availability. Some clients may be frequently offline or temporarily disconnected, which results in them receiving fewer model updates or improvements over time. To assess fairness in such temporally skewed participation settings, we must move beyond static or per-round metrics and instead evaluate the \emph{long-term utility} each client receives from the training process.
To this end, we define the cumulative utility received by each client up to round $T$ and normalize it by the client’s expected availability. This Lemma shows that under mild assumptions (stationary availability, bounded utility), this fairness criterion converges as training progresses.

\textbf{Theorem 1.}
Let \(u_k(T)\) be the cumulative utility accrued by client \(k\in[m]\) up to round \(T\), and let
\(\pi_k\in(0,1]\) be its long-run availability.
Define the availability-normalized utility vector
\begin{equation}
\begin{aligned}
\tilde u(T) \;=\; \Big(\frac{u_1(T)}{\pi_1},\dots,\frac{u_m(T)}{\pi_m}\Big),\\
\qquad
V_T \;:=\; \mathrm{Var}(\tilde u(T)) \;=\; \frac1m\sum_{k=1}^m\Big(\frac{u_k(T)}{\pi_k}-\bar u(T)\Big)^2,
\end{aligned}
\end{equation}
where \(\bar u(T)=\frac1m\sum_{k=1}^m u_k(T)/\pi_k\).

Consider two training schemes:
(i) \emph{vanilla} sampling/aggregation without availability compensation, producing \(w_{\mathrm{vanilla}}\);
(ii) an \emph{availability-aware, utility-compensated} scheme using inverse-availability correction, producing \(w_{\mathrm{fair}}\).
Under Assumptions~\textnormal{(A1)--(A4)} mentioned in appendix B, for each fixed horizon \(T\),
\begin{equation}
\mathbb{E}\!\left[V_T\big(w_{\mathrm{fair}}\big)\right]
\;\le\;
\mathbb{E}\!\left[V_T\big(w_{\mathrm{vanilla}}\big)\right]
\end{equation}
with strict inequality whenever availabilities are heterogeneous and utility increments have non-degenerate mean.

See the proof in appendix B. These guarantees characterize expected behavior under stochastic availability and are not intended to cover adversarial or fully bursty participation, which we leave as an open problem.

\textbf{Lemma 2.}
Let $N$ be the total number of clients. For client $k$ let $A_k(t)\in\{0,1\}$ be the availability indicator at round $t$ and assume the availability processes $\{A_k(t)\}_{t\ge 1}$ are stationary and ergodic with means $\pi_k=\mathbb{E}[A_k(t)]>0$. 
Let $\hat{\pi}_k(t)$ be an estimator of $\pi_k$ satisfying $\hat{\pi}_k(t)\to\pi_k$ as $t\to\infty$, and define inverse-availability weights
\begin{equation}
q_k(t)=\frac{1}{\hat{\pi}_k(t)}
\end{equation}
At each round the server selects $m$ clients by sampling among the currently \emph{available} clients with probability proportional to $q_k(t)$. Let $S_k(T)$ be the total number of times client $k$ is selected in $T$ rounds. Then, as $T\to\infty$,
\begin{equation}
\frac{\mathbb{E}[S_k(T)]}{T} \longrightarrow \frac{m}{N}, \qquad \forall k
\end{equation}
In other words, inverse-availability sampling equalizes the long-run selection frequency: each client is selected a fraction $m/N$ of the rounds in expectation.

\paragraph{Proof.}
At round $t$ the server's selection probability for client $k$, \emph{conditional on the availability vector} $A(t)=(A_1(t),\dots,A_N(t))$, equals
\begin{equation}
P_t(k \mid A(t)) \;=\; A_k(t)\cdot
\frac{q_k(t)}{\sum_{j=1}^N q_j(t) A_j(t)}
\end{equation}
because unavailable clients ($A_k(t)=0$) cannot be chosen and the probabilities are proportional to $q_j(t)$ over the available set.

Take expectations over the availability process. Using stationarity/ergodicity and the convergence $\hat\pi_j(t)\to\pi_j$, the random denominator satisfies the law of large numbers:
\begin{equation}
\sum_{j=1}^N q_j(t) A_j(t)
\;\xrightarrow{\;t\to\infty\;}\;
\sum_{j=1}^N q_j \pi_j
\;=\;
\sum_{j=1}^N \frac{1}{\pi_j}\pi_j
\;=\;
N
\end{equation}
almost surely, where $q_j:=\lim_{t\to\infty} q_j(t)=1/\pi_j$. Hence for large $t$,
\begin{equation}
P_t(k \mid A(t)) \approx A_k(t)\cdot \frac{q_k}{N}
\end{equation}
Taking the unconditional expectation and using $\mathbb{E}[A_k(t)]=\pi_k$ gives
\begin{equation}
\lim_{t\to\infty} \mathbb{E}[P_t(k \mid A(t))]
\;=\; \pi_k\cdot\frac{q_k}{N}
\;=\; \pi_k\cdot\frac{1/\pi_k}{N}
\;=\; \frac{1}{N}
\end{equation}
Thus the per-round expected selection probability for client $k$ tends to \(1/N\). If the server draws $m$ independent samples per round, the linearity of expectation yields a limiting expected number of selections per round equal to \(m/N\). Summing over \(T\) rounds and dividing by \(T\) gives
\begin{equation}
\lim_{T\to\infty}\frac{\mathbb{E}[S_k(T)]}{T}=\frac{m}{N}
\end{equation}
which holds for every client \(k\).

We discuss non-stationary availability using sliding-window fairness in Appendix C, demonstrating that cumulative utility parity degrades gracefully under non-stationary and correlated dropout patterns.

\textbf{Theorem 2.}
Let $A_k(t)\in\{0,1\}$ be client $k$'s availability indicator and assume the availability processes are stationary and ergodic with mean $\pi_k=\mathbb{E}[A_k(t)]\in(0,1]$. 
Define the missed-count
\begin{equation}
\mathrm{missed}_k(t) \;=\; \sum_{s=1}^{t-1} \bigl(1 - A_k(s)\bigr)
\end{equation}
and consider the reactive weight
\begin{equation}
p_k(t) \;=\; \frac{\alpha_k}{\pi_k + \epsilon}\cdot\bigl(1 + \lambda\,\mathrm{missed}_k(t)\bigr)
\end{equation}
for constants $\alpha_k>0$, $\lambda\ge 0$, and small $\epsilon>0$. Let the normalized (per-round) weight be
\begin{equation}
\widehat p_k(t) \;=\; \frac{p_k(t)}{\sum_{j=1}^N p_j(t)}
\end{equation}
Then under the stationarity assumption,
\begin{equation}
\mathbb{E}[\mathrm{missed}_k(t)] = (t-1)\,(1-\pi_k)
\end{equation}
and the normalized weight admits the asymptotic limit
\begin{equation}
\lim_{t\to\infty}\widehat p_k(t)
= 
\frac{\alpha_k \dfrac{1-\pi_k}{\pi_k}}{\sum_{j=1}^N \alpha_j \dfrac{1-\pi_j}{\pi_j}}
\qquad\text{provided }\lambda>0.
\end{equation}
If $\lambda=0$ then
\begin{equation}
\lim_{t\to\infty}\widehat p_k(t)
= 
\frac{\alpha_k/\pi_k}{\sum_{j=1}^N \alpha_j/\pi_j}
\end{equation}

The proof is discussed in appendix D. Afterall,
Lemma 2 focuses on the \textit{sampling stage}, demonstrating that using inverse-availability weighted sampling ensures each client is selected with equal expected frequency asymptotically, effectively compensating for differences in client availability. Theorem 2 considers the \textit{dynamic adjustment of participation weights} based on the history of missed rounds, where clients who have missed more rounds receive progressively higher weights in subsequent rounds.

\textbf{Lemma 3.}
Let $\mathcal{A}_t$ be the set of available clients and $\mathcal{M}_t$ the set of unavailable (missing) clients at training round $t$. Suppose the global objective function at round $t$ is defined with surrogate updates as:
\begin{equation}
F_{\text{global}}(w) = \sum_{k \in \mathcal{A}_t} q_k F_k(w) + \sum_{k' \in \mathcal{M}_t} \eta_{k'} \tilde{F}_{k'}(w)
\end{equation}
where $\tilde{F}_{k'}(w)$ is a surrogate estimate of the true loss $F_{k'}(w)$ for client $k'$; $q_k$ are the aggregation weights assigned to the available clients in round t, and $\eta_{k'}$ is the surrogate aggregation weight. Assume the surrogate error is uniformly bounded:
\begin{equation}
\| \tilde{F}_{k'}(w) - F_{k'}(w) \| \leq \epsilon, \quad \forall k' \in \mathcal{M}_t
\end{equation}
Then, the total deviation of the surrogate-corrected global loss from the full-data global loss is bounded as:
\begin{equation}
\left\| \sum_{k' \in \mathcal{M}_t} \eta_{k'} \left( \tilde{F}_{k'}(w) - F_{k'}(w) \right) \right\| \leq \epsilon \cdot \sum_{k' \in \mathcal{M}_t} \eta_{k'}
\end{equation}
\paragraph{Proof.}
Define the bias in the global loss due to surrogate error as:
\begin{equation}
\Delta(w) := \sum_{k' \in \mathcal{M}_t} \eta_{k'} \left( \tilde{F}_{k'}(w) - F_{k'}(w) \right)
\end{equation}
Then by the triangle inequality for vector norms:
\begin{equation}
\begin{aligned} 
\left\| \Delta(w) \right\| &= \left\| \sum_{k' \in \mathcal{M}_t} \eta_{k'} \left( \tilde{F}_{k'}(w) - F_{k'}(w) \right) \right\| \\
&\leq \sum_{k' \in \mathcal{M}_t} \left\| \eta_{k'} \left( \tilde{F}_{k'}(w) - F_{k'}(w) \right) \right\|
\end{aligned}
\end{equation}
Using the scalar factorization property of norms:
\begin{equation}
\leq \sum_{k' \in \mathcal{M}_t} \eta_{k'} \cdot \left\| \tilde{F}_{k'}(w) - F_{k'}(w) \right\|
\end{equation}
Since $\left\| \tilde{F}_{k'}(w) - F_{k'}(w) \right\| \leq \epsilon$ for all $k'$, we conclude:
\begin{equation}
\left\| \Delta(w) \right\| \leq \epsilon \cdot \sum_{k' \in \mathcal{M}_t} \eta_{k'}
\end{equation}

\textbf{Theorem 3.}
Fix a communication round \(t\). Let \(\mathcal{M}_t\subseteq [N]\) denote the set of clients whose contribution at round \(t\)
is replaced by a stale surrogate constructed from the last available update at time \(\tau_{k'}<t\).
Define the staleness \(\delta_{k'} := t-\tau_{k'}\in\mathbb{N}\).
Let \(F_k(w)\in\mathbb{R}^d\) denote the \emph{true} client signal used by the server (e.g., gradient \(\nabla f_k(w)\),
a control variate, or any vector-valued statistic), and let \(\tilde F_{k'}(w)\) be its surrogate for \(k'\in\mathcal{M}_t\).

See the proof in appendix E. Afterall,
Lemma 3 and Theorem 3 establish theoretical guarantees on the impact of using surrogate updates for unavailable clients in federated learning. Lemma 3 shows that when surrogate approximations of client losses are bounded by a uniform error $\epsilon$, the total deviation introduced in the global loss is linearly bounded by the aggregation weights assigned to the missing clients. This ensures that even when real updates are missing, the global objective does not drift arbitrarily far from the full-data objective, provided surrogates are sufficiently accurate and weighted conservatively. Theorem 3 extends this result by incorporating the notion of staleness.

\section{Empirical Evaluation}
\label{s:evaluation}

We now empirically validate the theoretical contributions of our availability-aware fairness framework in federated learning. We use a non-IID benchmark dataset widely used in FL studies: CIFAR-10 (image classification). For CIFAR-10, data is partitioned into clients such that each client only observes a subset of classes \cite{q-FFL, FedAvg, FedFV, fairfedcs2023}.

\textbf{Workload characteristics.}
CIFAR-10 occupies approximately 170 MB on disk and about 220 MB in memory when fully resident in \texttt{float32}. We use ResNet-18 and ResNet-34 models on containerized clients, with approximately 11.7M (45 MB) and 21.8M (85 MB) parameters, respectively. With batch size 32 and no gradient storage during inference, per-process memory usage is typically below 300 MB, allowing multiple concurrent clients and a coordinator to run comfortably within a few gigabytes of RAM. Training incurs higher memory usage due to activations and optimizer state; to accommodate this, all experiments are conducted on a single server with 64 GB RAM and modest local batch sizes. Although clients share the same underlying CIFAR-10 files via a host-mounted dataset, training and evaluation data are logically skewed per client. Each client constructs a non-IID local dataset by selecting a fixed subset of samples, restricted to a small number of labels (two labels per client in our experiments), inducing label-skewed data partitions. On the server side, evaluation is performed using the same non-IID scheme, ensuring that both training and per-client evaluation reflect heterogeneous.

\begin{figure}
  \centering
  \includegraphics[width=1.0\linewidth]{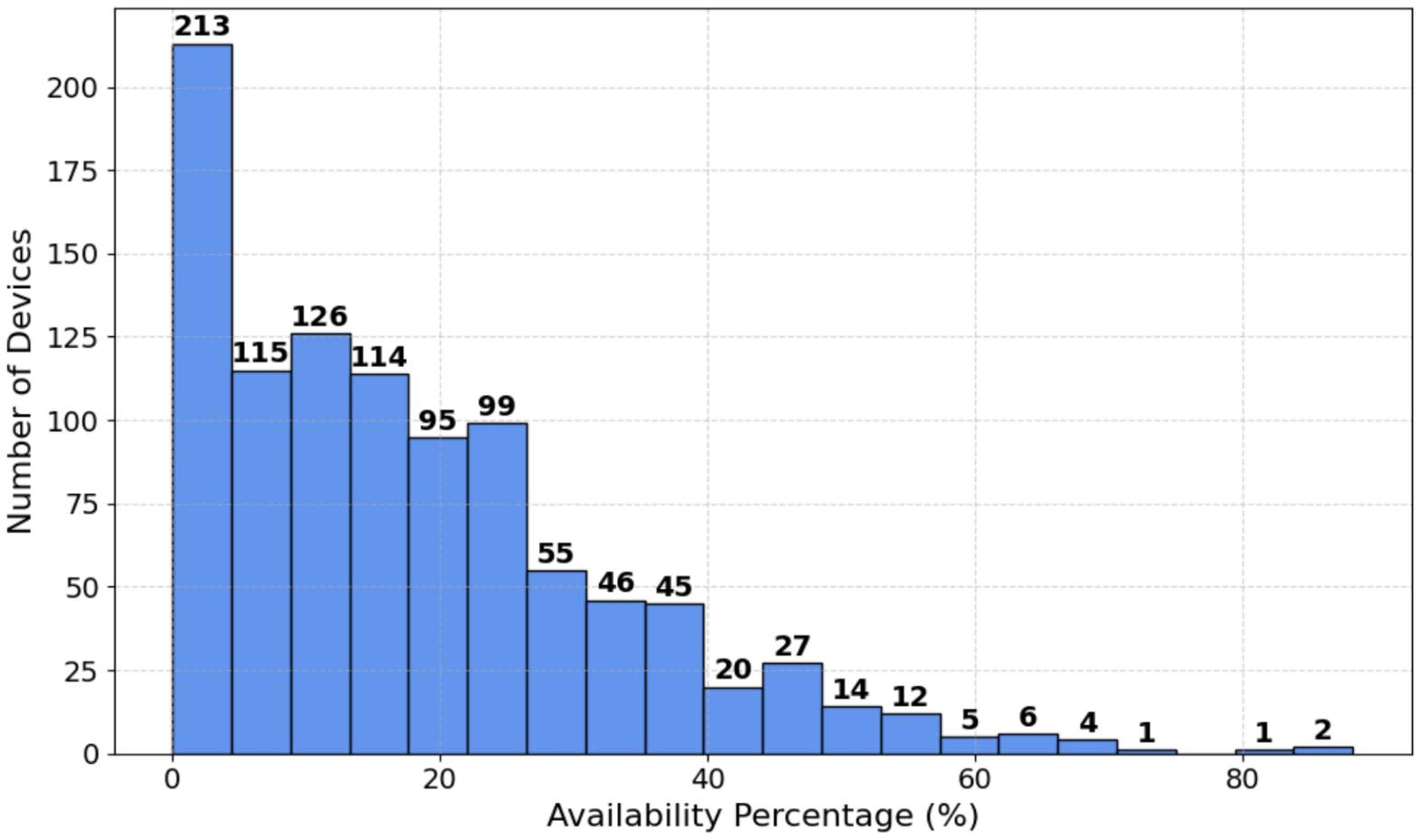}
\caption{\label{f:distribution}Distribution of device availability percentages
    for the trace data~\cite{yang2021characterizing}, where \emph{device
      availability percentage} is defined as the percentage of time between the
    first and last times a device was seen to be live and available to
    perform FL, i.e.,~was charging and connected to Wi-Fi. Out of 1000 devices in
    the trace, 213 were available for $<$5\% of the time, and over 60\% were available for less than half the time.}
\end{figure}

\textbf{Availability.} 
Our implementation is based on the real-world availability traces of mobile devices in an FL system published by~\cite{yang2021characterizing}. The distribution of device availability percentages indicates that most devices are available less than 40\% of the time (Figure~\ref{f:distribution}), where availability percentage is defined as the fraction of time (between the first and last observation of a device) during which it is in a state suitable for FL participation. 
The trace records timestamped events such as \emph{WiFi on/off} and \emph{battery charging on/off}. Following Google’s FL readiness definition~\cite{GoogleFL}, we treat a device as available when it is simultaneously charging and connected to WiFi. Unavailability therefore corresponds to common user-driven events—such as unplugging the phone, moving out of WiFi.

\subsection{Fairness via Temporal Utility Tracking}

To measure fairness under intermittent participation, we track each client’s cumulative utility and normalize it by their availability. This process involves several key steps. First, we log participation, which records whether a client was selected in each round—this acts as the binary availability indicator $A_k(t)$. We compute the change in client loss before and after inference (or local training), approximating $\Delta F_k(t)$, which represents the utility gained by the client in that round. We then accumulate utility per client as $u_k(t+1) \leftarrow u_k(t) + A_k(t) \cdot \Delta F_k(t)$. After each round, we estimate the long-term availability $\pi_k$ for each client based on how often they’ve been selected up to that point. Finally, we compute the normalized utility $\tilde{u}_k = u_k / \pi_k$, and track the variance across clients for fairness (so-called Fairness Variance). 
Lemma~1 provides a fairness target: if the variance of $\tilde{u}_k$ is bounded, then representation parity is achieved in the long term. Theorem-1 gives a concrete mechanism to move toward this goal: by sampling clients with probabilities inversely proportional to their availability (i.e., $\propto 1/\pi_k$), we equalize the effective number of participation opportunities across clients. This compensates for skewed participation patterns and gives low-availability clients more chances to contribute and accumulate utility. 


\begin{algorithm}[t]
\caption{Server Orchestration: Fair/Vanilla FL with Logging}
\label{alg:server}
\small
\begin{algorithmic}[1]
\REQUIRE Total rounds $T$, clients-per-round $m$, initial model $w^{(0)}$; availability indicators $\{A_k(t)\}$
\STATE Initialize registry $\mathcal{K}\leftarrow\emptyset$ 
\STATE Initialize logs $\{u_k^{\text{fair}}\},\{u_k^{\text{van}}\}$ and counts $\{s_k^{\text{fair}}\},\{s_k^{\text{van}}\}$
\STATE Wait until enough clients register; set $w_{\text{fair}}\!\leftarrow\!w^{(0)}$, $w_{\text{van}}\!\leftarrow\!w^{(0)}$
\FOR{$t=1$ {\bf to} $T$}
    \STATE Observe available clients $\mathcal{A}_t \subseteq \mathcal{K}$ from $A_k(t)$
    \STATE Update $\hat{\pi}_k(t)\leftarrow \frac{1}{t}\sum_{\tau=1}^{t} A_k(\tau)$ for all $k\in\mathcal{K}$; log $|\mathcal{C}_t|$
    \COMMENT{\textbf{Fair selection branch (ours)}}
    \STATE $(\mathcal{S}^{\text{fair}}_t, \textsc{VarU}_t, \textsc{Bias}_t)\leftarrow \textsc{SelectFair}(\mathcal{A}_t,w_{\text{fair}},\hat{\pi}(t),u^{\text{fair}},s^{\text{fair}},m)$
    \STATE $w_{\text{fair}}\leftarrow \textsc{FederatedRound}(\mathcal{S}^{\text{fair}}_t, w_{\text{fair}})$
    \STATE $\textsc{Acc}^{\text{fair}}_t \leftarrow \textsc{Evaluate}(w_{\text{fair}})$
    \COMMENT{\textbf{Vanilla (random) baseline}}
    \STATE $\mathcal{S}^{\text{van}}_t \leftarrow \textsc{SelectRandom}(\mathcal{A}_t,m)$
    \STATE $w_{\text{van}}\leftarrow \textsc{FederatedRound}(\mathcal{S}^{\text{van}}_t, w_{\text{van}})$
    \STATE $\textsc{Acc}^{\text{van}}_t \leftarrow \textsc{Evaluate}(w_{\text{van}})$
    \COMMENT{\textbf{Per-round + per-client logging}}
    \STATE $\textsc{Row}^{\text{fair}}_t \leftarrow \textsc{ComputeMetrics}(w_{\text{fair}},u^{\text{fair}},\hat{\pi}(t),s^{\text{fair}},t)$
    \STATE $\textsc{Row}^{\text{van}}_t \leftarrow \textsc{ComputeMetrics}(w_{\text{van}},u^{\text{van}},\hat{\pi}(t),s^{\text{van}},t)$
    \STATE Append $\textsc{Row}^{\text{fair}}_t$ and $\textsc{Row}^{\text{van}}_t$ to \texttt{metrics\_log.csv}
\ENDFOR
\end{algorithmic}
\end{algorithm}

\begin{algorithm}[t]
\caption{Client Procedure: Register, Train Locally, Return Update}
\label{alg:client}
\small
\begin{algorithmic}[1]
\REQUIRE Local data loader $\mathcal{D}_k$, local epochs $E$, optimizer $\textsc{Opt}$, mixing factor $\alpha\in[0,1]$ 
\STATE Initialize local model parameters $w_k$; register client $k$ with the server
\WHILE{server requests training}
    \STATE Receive global weights $w^{(t)}$ from server
    \STATE \COMMENT{we do a warm-start used in deployment; not required for theoretical analysis}
    \STATE $w_k \leftarrow (1-\alpha)\,w_k + \alpha\,w^{(t)}$ \hfill (parameter-wise interpolation)
    \FOR{$e=1$ {\bf to} $E$}
        \FOR{mini-batch $b \sim \mathcal{D}_k$}
            \STATE $g \leftarrow \nabla \ell(w_k; b)$
            \STATE $w_k \leftarrow \textsc{Opt}(w_k, g)$
        \ENDFOR
    \ENDFOR
    \STATE Send updated weights $w_k$ (or failure signal) back to server
    \STATE Advance availability trace / telemetry and update local stats (latency, health)
\ENDWHILE
\end{algorithmic}
\end{algorithm}

\begin{figure}[h]
\centering \includegraphics[width=0.9\linewidth]{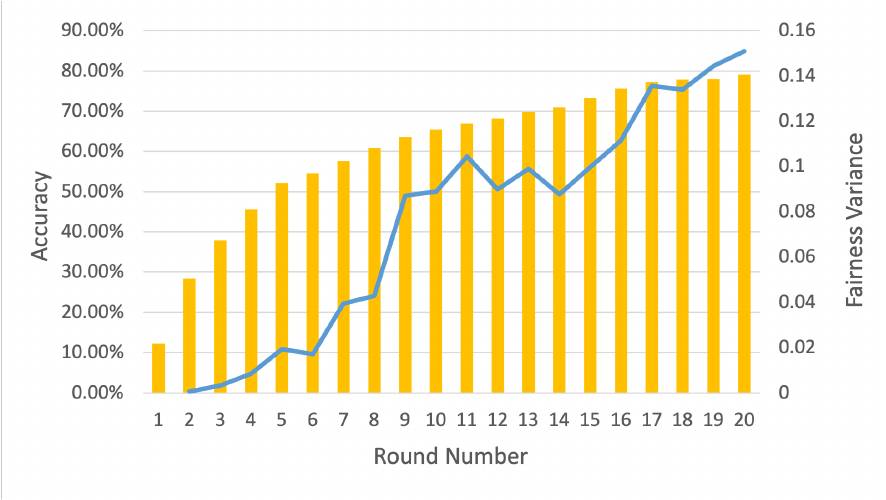}
\caption{\label{random-sampling}Accuracy and fairness variance versus round number using random sampling.}
\end{figure}

\begin{figure}[h]
\centering     \includegraphics[width=0.9\linewidth]{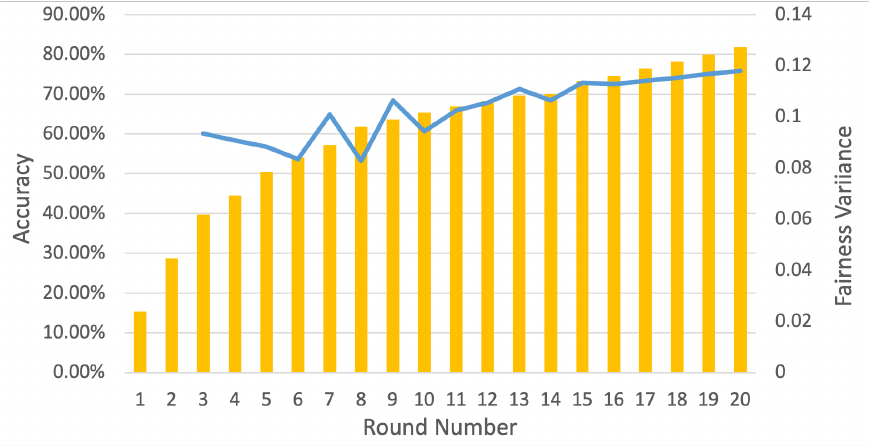}
\caption{\label{inverse-sampling}Accuracy and fairness variance versus round number using inverse-availability × missed-round reweighting ($\lambda=0.7$).}
\end{figure}

\subsection{Adaptive Sampling and Participation Balance}

Under our inverse-availability sampling and reactive reweighting (Lemma 2 and Theorem 2), participation frequencies become balanced over time, even under skewed $\pi_k$. 
At the core of our approach is the observation that random sampling fails to account for persistent disparities in availability (increasing variance in Figure~\ref{random-sampling}), leading to over-selection of frequently available clients and underrepresentation of low-availability ones. To correct this, we adopt an inverse-availability sampling scheme (Figure~\ref{inverse-sampling}) where the probability of selecting client $k$ at round $t$ is proportional to the inverse of its estimated long-term availability, i.e., $\textit{Inverse-Availability × Missed-Round Reweighting}$. 
At each round, we compute these scores for all currently available clients and select the top-$K$ clients with the highest scores. Throughout training, we maintain per-client participation statistics to evaluate fairness in sampling. Let $s_k(T)$ be the number of times client $k$ has been selected up to round $T$. We track the deviation of each client’s participation frequency from the uniform target and compute the standard deviation as a measure of selection imbalance.

\begin{table}[ht]
\centering
\scriptsize
\caption{\label{Surrogate-Tab}Accuracy, fairness variance, and surrogate contribution over 20 rounds.}
\begin{tabular}{|p{1cm}|p{1.5cm}|p{1.5cm}|p{2.2cm}|}
\hline
\textbf{Round} & \textbf{Accuracy (\%)} & \textbf{Fairness Variance} & \textbf{Surrogate Contribution} \\
\hline
1  & 14.37 & ---     & 5.45 \\
2  & 26.25 & 0.0003  & 8.24 \\
3  & 36.09 & 0.0072  & 11.12 \\
4  & 44.22 & 0.1099  & 14.12 \\
5  & 46.41 & 0.2604  & 14.12 \\
6  & 50.78 & 0.2490  & 14.12 \\
7  & 53.28 & 0.2283  & 16.94 \\
8  & 56.88 & 0.2348  & 16.94 \\
9  & 58.13 & 0.2054  & 19.95 \\
10 & 59.84 & 0.1689  & 22.99 \\
11 & 61.56 & 0.1406  & 22.99 \\
12 & 64.22 & 0.1294  & 22.99 \\
13 & 65.47 & 0.1155  & 22.99 \\
14 & 66.88 & 0.1168  & 22.99 \\
15 & 68.44 & 0.1090  & 22.99 \\
16 & 70.78 & 0.1087  & 22.99 \\
17 & 72.34 & 0.0984  & 22.99 \\
18 & 73.28 & 0.1086  & 22.99 \\
19 & 74.22 & 0.1146  & 22.99 \\
20 & 74.38 & 0.1140  & 22.99 \\
\hline
\end{tabular}
\label{tab:surrogate_results}
\end{table}

\begin{table*}[t]
\centering
\caption{Comparison of performance fairness and cumulative participation fairness under intermittent client participation.
Higher is better ($\uparrow$) and lower is better ($\downarrow$).}
\label{tab:fairness-comparison}
\begin{tabular}{lcccccc}
\toprule
Method 
& Avg Acc $\uparrow$ 
& Jain (Acc) $\uparrow$ 
& Utility CV $\downarrow$ 
& Jain (Utility) $\uparrow$  
& Sel.\ Gap $\downarrow$ 
& Gini $\downarrow$ \\
\midrule
q-FFL 
& 60.1
& 0.72
& 0.64 
& 0.42
& 0.80 
& 0.35 \\

PHP-FL 
& 67.71 
& 0.80
& 0.42 
& 0.78 
& 0.52
& 0.20 \\

\textbf{Ours (no surrogate)}
& 80.43
& 0.975
& 0.28
& 0.88
& 0.31
& 0.04 \\

\textbf{Ours (with surrogate)} 
& \textbf{80.43} 
& \textbf{0.975} 
& \textbf{0.19} 
& \textbf{0.94} 
& \textbf{0.31} 
& \textbf{0.04} \\
\bottomrule
\end{tabular}

\vspace{0.5em}
\footnotesize
\textbf{Notes.}
Jain (Acc) is computed over per-client test accuracies.
Utility CV and Jain (Utility) are computed over availability-normalized cumulative utilities.
Selection Gap denotes the $\ell_1$ deviation from uniform client selection normalized by T.
\end{table*}

\subsection{Impact of Surrogate Updates}
While inverse-availability sampling increases selection priority for infrequently participating clients, it does not preserve representation when clients are temporarily missing, as their data distributions are omitted from the current update. 
This effect is amplified under correlated dropout, leading to erosion of rare class representations. 
To mitigate this, we incorporate surrogate updates. Also to validate Lemma 3 and Theorm 3, we simulate client dropout and apply surrogates trained from previous checkpoints. 
For clients that are unavailable in a given round, we estimate their potential contribution using a surrogate constructed from their last observed model state. 
Specifically, when a client $k$ is missing at round $t$, we retrieve its most recent model parameters and define the staleness $\delta_k(t)$ as the number of rounds since its last participation. 
The surrogate contribution is weighted by an exponentially decaying reliability factor $\eta_k(t)=\eta_0 \exp(-\lambda \delta_k(t))$, which captures increasing uncertainty with staleness. 
The surrogate model is evaluated on the client’s fixed local data distribution using the same loss function as active clients, and its downweighted contribution is incorporated into cumulative utility tracking without introducing stale parameter updates into the optimization process.
Surrogate updates are not required for the fairness guarantees in this work, and all theoretical results hold independently of their use.

As seen in Table~\ref{Surrogate-Tab}, in the early rounds (1--4), the model accuracy improves rapidly from 14.37\% to 44.22\% as more clients begin participating and surrogate updates help fill gaps caused by unavailability. During this period, the surrogate contribution grows steadily from 5.45 to 14.12, reflecting active reliance on past updates from dropped clients. This supports the theoretical behavior described in Theorem 3, where surrogate contributions are weighted by staleness and provide continuity in learning. Fairness variance, initially near zero, begins to rise, indicating that while surrogate updates help performance, they introduce disparity in normalized utility due to unequal access and contribution.
From rounds 5 onward, the accuracy continues to improve, eventually reaching to around 74\%. Meanwhile, fairness variance peaks in rounds 5--6 and then gradually decreases, showing that the inverse-availability sampling and reactive reweighting mechanisms are balancing out long-term participation opportunities. Notably, the surrogate contribution reaches a ceiling at 22.99 and remains constant for the rest, suggesting that the surrogate pool stabilizes as the same clients remain unavailable.

\section{Baseline comparison}
\label{s:baseline}

\subsection{Metrics for evaluation}
To address fairness under intermittent client participation, we evaluate methods along two
complementary axes:
(i) \emph{performance fairness}, which captures disparities in predictive quality across clients, and
(ii) \emph{cumulative participation fairness}, which captures how equitably clients are represented and benefit over the entire training process.

\vspace{0.5em}
\noindent\textbf{(i) Performance-based metrics.}
We report standard performance fairness metrics commonly used:
\begin{itemize}
  \setlength{\itemsep}{0pt}
  \setlength{\parskip}{0pt}
  \setlength{\parsep}{0pt}
    \item \textbf{Average Accuracy (Avg Acc)}
    \item \textbf{Jain's Fairness Index (Accuracy)}:
    \begin{equation}
    J(a) = \frac{\left(\sum_{k=1}^{N} a_k\right)^2}{N \sum_{k=1}^{N} a_k^2}
    \end{equation}
    where $a_k$ denotes the accuracy of client $k$.
\end{itemize}

\noindent\textbf{(ii) Cumulative participation metrics.}

\textit{Availability-normalized cumulative utility.}
Let $\Delta u_k(t)$ denote the utility gained by client $k$ at round $t$.
The cumulative utility of client $k$ is defined in Eq. 1.

\begin{itemize}
  \setlength{\itemsep}{0pt}
  \setlength{\parskip}{0pt}
  \setlength{\parsep}{0pt}
     \item \textbf{Utility Coefficient of Variation (Utility CV)}:
    \begin{equation}
    \mathrm{CV}(\tilde{u}) = \frac{\sigma(\tilde{u})}{\mu(\tilde{u})+\epsilon},
    \end{equation}
     which measures relative dispersion of normalized cumulative utility, where $\mu(\tilde{u})$ and $\sigma(\tilde{u})$ denote the mean and standard deviation of $\{\tilde{u}_k\}_{k=1}^N$ across clients, and $\epsilon>0$ is a small number ensuring numerical stability.

    \item \textbf{Jain's Fairness Index (Utility)}: $J(\tilde{u})$, which captures equality of cumulative benefit across clients.
\end{itemize}

Lower Utility CV and higher Jain index indicate stronger cumulative fairness.

\textit{Participation parity.}
Let $S_k$ denote the total number of rounds in which client $k$ is selected for training.
Assuming $m$ clients are selected per round, the ideal fair selection share for each client is $T/N$ rounds.
We quantify deviations from this ideal using:
\begin{itemize}
  \setlength{\itemsep}{0pt}
  \setlength{\parskip}{0pt}
  \setlength{\parsep}{0pt}
    \item \textbf{Selection Gap ($\ell_1$)}:
    \begin{equation}
    \text{SelGap}_{\ell_1} =
    \frac{1}{T}\sum_{k=1}^{N}
    \left|
    \frac{S_k}{m} - \frac{T}{N}
    \right|
    \end{equation}
    \item \textbf{Gini Coefficient of Selection Counts}, which measures inequality in client participation using standard metric.
\end{itemize}

\subsection{Baseline evaluations}
To comprehensively evaluate fairness under intermittent participation, we report both performance-based and allocation-based metrics. Model performance is assessed using the average accuracy across clients and the accuracy variance. 
While such metrics are standard in fair federated learning and are used by prior work such as loss-reweighting and update-based methods, they do not capture disparities arising from unequal participation over time.
To address this limitation, we introduce temporal fairness metrics that operate on cumulative quantities. 
Specifically, we compute availability-normalized cumulative utility for each client, defined as the total utility accrued over training normalized by the client’s empirical availability, and measure its dispersion using the coefficient of variation (CV) and Jain’s fairness index, which respectively quantify relative spread and equality of long-term benefit. 
In addition, to directly evaluate fairness in the allocation of training opportunities, we measure representation parity using the selection gap, defined as the deviation between the empirical selection frequency of a client and its ideal fair share.

When comparing our method to q-FFL and PHP-FL, we evaluate performance using per-client accuracy rather than only global accuracy, which aggregates performance across all data points and can therefore mask systematic disparities among clients, particularly under intermittent participation. 
For baseline methods that do not explicitly track cumulative utility, we compute per-client utility retrospectively using a consistent evaluation protocol. 
Specifically, per-round utility increment $\Delta u_k(t)$ is measured as the change in per-client accuracy between consecutive rounds. 
Cumulative utility is then obtained by summing these increments over training and normalizing by the client’s empirical availability.

Table~\ref{tab:fairness-comparison} reports the performance and cumulative fairness metrics under intermittent participation for a federation of 100 clients over 50 training rounds. 
While prior methods such as q-FFL and PHP-FL improve loss-based fairness among participating clients, they exhibit substantial dispersion in availability-normalized cumulative utility (Utility CV = 0.64 and 0.42, respectively). 
In contrast, our method achieves significantly lower utility dispersion (Utility CV = 0.19) and maintains bounded representation disparity throughout training. 
Importantly, this improvement in cumulative participation parity does not come at the expense of model quality: our approach attains near-perfect accuracy fairness (Jain = 0.975), while also achieving the lowest selection gap and Gini coefficient with or without of using surrogate.

\section{Conclusion}
In this work, we studied fairness in federated learning under the realistic and often overlooked setting of intermittent client participation. We showed that fairness notions based solely on per-round loss or accuracy implicitly assume consistent participation and can therefore mask long-term under-representation of intermittently available clients. To address this gap, we introduced a resource-allocation perspective on fairness, formalized through availability-normalized cumulative utility, which captures whether clients receive comparable benefit per participation opportunity over time.
We provided theoretical guarantees showing contraction of cumulative utility disparity under intermittent availability, and demonstrated empirically that our approach achieves improved representation parity and reduced fairness dispersion without sacrificing predictive performance. Together, these results emphasize that fairness in federated learning is inherently temporal and participation-dependent, and cannot be fully characterized by per-round accuracy parity alone.

\section*{Impact Statement}
This paper presents work whose primary goal is to advance the theoretical and algorithmic foundations of fairness in federated learning under intermittent client participation. The methods developed in this work aim to improve equitable allocation of learning benefits across clients in distributed systems where participation is irregular and constrained by user behavior or system dynamics.

Potential positive societal impacts include more equitable treatment of underrepresented or intermittently connected devices, which may help reduce performance disparities in real-world federated deployments such as mobile, edge, or privacy-sensitive learning systems. By explicitly accounting for participation opportunity over time, the proposed framework may also support more transparent and accountable fairness analysis in decentralized machine learning.

The proposed methods do not introduce new privacy risks beyond those inherent to standard federated learning settings, as they rely on aggregate, availability-level statistics and do not require access to raw client data. 
Overall, we do not foresee significant negative societal consequences arising uniquely from this work beyond those already associated with federated learning systems. The contribution is intended as a conceptual and methodological step toward more robust and principled fairness mechanisms in distributed machine learning.


\bibliography{References}
\bibliographystyle{icml2026}

\newpage
\appendix
\onecolumn

\section{Quantitative bound on normalized-utility deviation under inverse-availability sampling}
Let there be $N$ clients. For client $k$ define the availability indicator $A_k(t)\in\{0,1\}$ with long-term mean $\pi_k=\mathbb{E}[A_k(t)]>0$.  
Let $S_k(t)\in\{0,1\}$ be the indicator that client $k$ is selected by the server at round $t$ (so a client contributes only when $A_k(t)S_k(t)=1$).  Define the per-round marginal utility $\Delta F_k(t)$ and assume it is bounded and has a constant mean
\begin{equation}
0 \le \Delta F_k(t) \le M,\qquad \mu_k := \mathbb{E}[\Delta F_k(t)]
\end{equation}
Let the cumulative utility up to $T$ be
\begin{equation}
u_k(T)=\sum_{t=1}^T A_k(t) S_k(t)\Delta F_k(t)
\end{equation}
Assume the server uses inverse-availability sampling with weights $q_j(t)=1/\hat\pi_j(t)$ and the estimators satisfy $\hat\pi_j(t)\to\pi_j$ so that (as in Lemma~2)
\begin{equation}
\lim_{t\to\infty}\Pr\big(S_k(t)=1\mid A_k(t)=1\big)=\frac{1/\pi_k}{\sum_{j=1}^N 1/\pi_j} =: \frac{1/\pi_k}{C}
\end{equation}
with $C:=\sum_{j=1}^N 1/\pi_j$. Then for every client $k$ and every horizon $T$ (asymptotically in the estimator convergence),
\begin{equation}
\mathbb{E}\!\left[\frac{u_k(T)}{\pi_k}\right]
= \frac{T}{C\pi_k}\,\mu_k,
\end{equation}
and the mean normalized utility across clients is
\begin{equation}
\bar u(T) \;:=\; \frac{1}{N}\sum_{j=1}^N \mathbb{E}\!\left[\frac{u_j(T)}{\pi_j}\right]
= \frac{T}{CN}\sum_{j=1}^N \frac{\mu_j}{\pi_j}
\end{equation}
Consequently the deviation satisfies the exact identity
\begin{equation}
\Biggl|\mathbb{E}\!\left[\frac{u_k(T)}{\pi_k}\right] - \bar u(T)\Biggr|
= \frac{T}{C}\Biggl|\frac{\mu_k}{\pi_k} - \frac{1}{N}\sum_{j=1}^N\frac{\mu_j}{\pi_j}\Biggr|
\end{equation}
In particular, using $\mu_j\in[0,M]$ and $\pi_j\ge \pi_{\min}>0$, we obtain the uniform bound
\begin{equation}
\Biggl|\mathbb{E}\!\left[\frac{u_k(T)}{\pi_k}\right] - \bar u(T)\Biggr|
\le \frac{2T M}{C\,\pi_{\min}}
\end{equation}

Condition on availability and selection. For a fixed round $t$,
\begin{equation}
\begin{aligned}
\mathbb{E}\big[A_k(t)S_k(t)\Delta F_k(t)\big]
&= \Pr(A_k(t)=1)\Pr(S_k(t)=1\mid \\
&A_k(t)=1)\,\mathbb{E}[\Delta F_k(t)]
\end{aligned}
\end{equation}
where we used that $\Delta F_k(t)$ has mean $\mu_k$ (and is bounded). By the inverse-availability sampling limit,
\begin{equation}
\Pr(S_k(t)=1\mid A_k(t)=1)\xrightarrow[t\to\infty]{}\frac{1/\pi_k}{C}
\end{equation}
Hence for large $t$ (and in the asymptotic estimator regime) the per-round expected contribution equals
\begin{equation}
\Pr(A_k(t)=1)\cdot\frac{1/\pi_k}{C}\cdot\mu_k
= \pi_k\cdot\frac{1/\pi_k}{C}\cdot\mu_k
= \frac{1}{C}\mu_k
\end{equation}
Summing over $t=1,\dots,T$ and dividing by $\pi_k$ gives the closed form
\begin{equation}
\mathbb{E}\!\left[\frac{u_k(T)}{\pi_k}\right] = \frac{T}{C\pi_k}\,\mu_k
\end{equation}
Averaging across $k$ yields the stated $\bar u(T)$. Subtracting the two expressions gives the exact deviation identity, and bounding $0\le\mu_j\le M$ and $\pi_j\ge\pi_{\min}>0$ yields the stated uniform bound via elementary inequalities:
\begin{equation}
\Biggl|\frac{\mu_k}{\pi_k}-\frac{1}{N}\sum_j\frac{\mu_j}{\pi_j}\Biggr|
\le \frac{M}{\pi_k}+\frac{1}{N}\sum_j\frac{M}{\pi_j}
\le \frac{2M}{\pi_{\min}}
\end{equation}
and multiplying by $T/C$ completes the proof.

\section{Variance reduction under inverse-availability compensation}
\textbf{Theorem 1.}
Let \(u_k(T)\) be the cumulative utility accrued by client \(k\in[m]\) up to round \(T\), and let
\(\pi_k\in(0,1]\) be its long-run availability.
Define the availability-normalized utility vector
\begin{equation}
\tilde u(T) \;=\; \Big(\frac{u_1(T)}{\pi_1},\dots,\frac{u_m(T)}{\pi_m}\Big),
\qquad
V_T \;:=\; \mathrm{Var}(\tilde u(T)) \;=\; \frac1m\sum_{k=1}^m\Big(\frac{u_k(T)}{\pi_k}-\bar u(T)\Big)^2
\end{equation}
where \(\bar u(T)=\frac1m\sum_{k=1}^m u_k(T)/\pi_k\).

Consider two training schemes:
(i) \emph{vanilla} sampling/aggregation without availability compensation, producing \(w_{\mathrm{vanilla}}\);
(ii) an \emph{availability-aware, utility-compensated} scheme using inverse-availability correction, producing \(w_{\mathrm{fair}}\).
Under Assumptions~\textnormal{(A1)--(A4)} below, for each fixed horizon \(T\),
\begin{equation}
\mathbb{E}\!\left[V_T\big(w_{\mathrm{fair}}\big)\right]
\;\le\;
\mathbb{E}\!\left[V_T\big(w_{\mathrm{vanilla}}\big)\right]
\end{equation}
with strict inequality whenever availabilities are heterogeneous and utility increments have non-degenerate mean
(precise condition in Step~3).

\paragraph{Assumptions.}
Let \(A_k(t)\in\{0,1\}\) indicate availability of client \(k\) at round \(t\), and let \(S_k(t)\in\{0,1\}\) indicate
whether client \(k\) is \emph{selected and participates} at round \(t\).
\begin{itemize}
\item[(A1)] \emph{Stationary availability}: For each \(k\), \(\{A_k(t)\}_{t\ge1}\) is stationary with
\(\mathbb{E}[A_k(t)]=\pi_k\).
\item[(A2)] \emph{Conditional sampling rule}: Conditional on availability at round \(t\),
the selection mechanism chooses a subset of available clients.
Write \(p_k^{\mathrm{van}} := \mathbb{P}(S_k(t)=1\mid A_k(t)=1)\) for vanilla and
\(p_k^{\mathrm{fair}} := \mathbb{P}(S_k(t)=1\mid A_k(t)=1)\) for fair.
\item[(A3)] \emph{Inverse-availability compensation}: The fair scheme uses inverse-availability correction so that
\begin{equation}
\label{eq:invpi}
\frac{\mathbb{E}[S_k(t)]}{\pi_k} = \rho
\quad \text{for all } k,
\end{equation}
for some constant \(\rho\in(0,1]\) determined by the per-round budget (expected number of participating clients).
Equivalently, since \(\mathbb{E}[S_k(t)]=\pi_k p_k^{\mathrm{fair}}\), this is \(p_k^{\mathrm{fair}} \equiv \rho\),
i.e., \emph{conditional on being available}, every client is selected with the same probability.
\item[(A4)] \emph{Bounded utility increments}: If client \(k\) participates at round \(t\), it accrues utility increment
\(\Delta u_k(t)\), and \(u_k(T)=\sum_{t=1}^T S_k(t)\Delta u_k(t)\).
Assume \(|\Delta u_k(t)|\le B\) almost surely and
\(\mu_k := \mathbb{E}[\Delta u_k(t)]\) exists (finite).
\end{itemize}

\textbf{Proof.}
Fix \(T\). Define the participation count \(N_k(T):=\sum_{t=1}^T S_k(t)\).
Then
\begin{equation}
\label{eq:uk-decomp}
\frac{u_k(T)}{\pi_k}
\;=\;
\frac{1}{\pi_k}\sum_{t=1}^T S_k(t)\Delta u_k(t)
\;=\;
\underbrace{\frac{N_k(T)}{\pi_k}\mu_k}_{\text{systematic term}}
\;+\;
\underbrace{\frac{1}{\pi_k}\sum_{t=1}^T S_k(t)\big(\Delta u_k(t)-\mu_k\big)}_{\text{noise term}}
\end{equation}
Let \(X_k := u_k(T)/\pi_k\), and denote \(\bar X := \frac1m\sum_{k=1}^m X_k\).
We will upper-bound \(\mathbb{E}\big[\mathrm{Var}(X)\big]\) under each scheme and compare.

\medskip
\noindent
\textbf{Step 1: A variance decomposition across clients.}
Using \(\mathrm{Var}(X)=\frac1m\sum_{k=1}^m (X_k-\bar X)^2\), we apply
\begin{equation}
(X_k-\bar X)^2
\;\le\;
2\Big( \big(\tfrac{N_k(T)}{\pi_k}\mu_k - \overline{\tfrac{N}{\pi}\mu}\big)^2
      + \big(\eta_k-\bar\eta\big)^2 \Big),
\end{equation}
where \(\eta_k := \frac{1}{\pi_k}\sum_{t=1}^T S_k(t)(\Delta u_k(t)-\mu_k)\) is the noise term in~\eqref{eq:uk-decomp}.
Taking expectation and averaging over \(k\) yields
\begin{equation}
\label{eq:var-split}
\mathbb{E}\!\left[\mathrm{Var}(X)\right]
\;\le\;
2\,\mathbb{E}\!\left[\mathrm{Var}\!\Big(\tfrac{N(T)}{\pi}\mu\Big)\right]
\;+\;
2\,\mathbb{E}\!\left[\mathrm{Var}(\eta)\right]
\end{equation}
Thus, the dispersion of normalized utilities is controlled by
(i) dispersion of \(\tfrac{N_k(T)}{\pi_k}\mu_k\) across clients and
(ii) dispersion of the centered noise \(\eta_k\) across clients.

\medskip
\noindent
\textbf{Step 2: The compensation scheme equalizes normalized participation counts.}
Under the fair scheme, Assumption~(A3) implies \(\mathbb{E}[S_k(t)]/\pi_k=\rho\) for all \(k\), hence
\begin{equation}
\label{eq:fair-count-mean}
\mathbb{E}\!\left[\frac{N_k(T)}{\pi_k}\right]
\;=\;
\frac{1}{\pi_k}\sum_{t=1}^T \mathbb{E}[S_k(t)]
\;=\;
\frac{T\,\mathbb{E}[S_k(1)]}{\pi_k}
\;=\;
T\rho
\quad\text{for all }k
\end{equation}
So, under fair training, the \emph{mean} of the normalized count \(N_k(T)/\pi_k\) is identical across clients.

By contrast, under vanilla sampling, there is generally no reason for \(p_k^{\mathrm{van}}\) to be constant in \(k\),
so
\begin{equation}
\mathbb{E}\!\left[\frac{N_k(T)}{\pi_k}\right]
=
T\,p_k^{\mathrm{van}}
\end{equation}
which varies with \(k\) unless \(p_k^{\mathrm{van}}\) happens to be uniform.
This is exactly the participation skew that inverse-availability correction is designed to remove.

\medskip
\noindent
\textbf{Step 3: Dispersion gap in the systematic term.}
Assume for simplicity of exposition that mean utilities are not adversarially heterogeneous, e.g.,
\(\mu_k\equiv \mu>0\) (or more generally \(\mu_k\) are bounded and weakly varying; see remark at end).
Then
\begin{equation}
\mathrm{Var}\!\Big(\tfrac{N(T)}{\pi}\mu\Big)
=\mu^2\,\mathrm{Var}\!\Big(\tfrac{N(T)}{\pi}\Big)
\end{equation}
Using~\eqref{eq:fair-count-mean}, the fair scheme centers all \(\tfrac{N_k(T)}{\pi_k}\) at the same mean \(T\rho\),
whereas vanilla centers \(\tfrac{N_k(T)}{\pi_k}\) at \(T p_k^{\mathrm{van}}\), which is non-constant across \(k\).
Hence, purely at the level of across-client means,
\begin{equation}
\label{eq:mean-var-gap}
\mathrm{Var}\!\Big(\mathbb{E}\big[\tfrac{N(T)}{\pi}\big]\Big)
=
\mathrm{Var}\big(T\rho,\dots,T\rho\big)
=0
\quad\text{(fair)},
\end{equation}
while
\begin{equation}
\mathrm{Var}\!\Big(\mathbb{E}\big[\tfrac{N(T)}{\pi}\big]\Big)
=
\mathrm{Var}\big(Tp_1^{\mathrm{van}},\dots,Tp_m^{\mathrm{van}}\big)
>0
\quad\text{(vanilla)}
\end{equation}
whenever \(p_k^{\mathrm{van}}\) is not identical across clients.
This already yields a strict separation in the systematic dispersion term unless vanilla is accidentally uniform.

More formally, apply the variance identity
\(\mathrm{Var}(Z)=\mathbb{E}[\mathrm{Var}(Z\mid\mathcal{G})]+\mathrm{Var}(\mathbb{E}[Z\mid\mathcal{G}])\)
with \(\mathcal{G}\) the \(\sigma\)-field generated by the scheme parameters.
Taking \(Z_k=\tfrac{N_k(T)}{\pi_k}\) and measuring variance across \(k\),
the second term is exactly the across-client variance of the means, which is \(0\) for fair and positive for vanilla
when \(p_k^{\mathrm{van}}\) is heterogeneous. This implies
\begin{equation}
\label{eq:sys-gap}
\mathbb{E}\!\left[\mathrm{Var}\!\Big(\tfrac{N(T)}{\pi}\mu\Big)\right]_{\mathrm{fair}}
\;<\;
\mathbb{E}\!\left[\mathrm{Var}\!\Big(\tfrac{N(T)}{\pi}\mu\Big)\right]_{\mathrm{vanilla}}
\end{equation}
whenever \(\mu>0\) and \(p_k^{\mathrm{van}}\) is not constant.

\medskip
\noindent
\textbf{Step 4: The noise term is not amplified by inverse-availability correction.}
We bound \(\mathbb{E}[\mathrm{Var}(\eta)]\) uniformly using bounded increments.
Conditioned on the selection indicators \(\{S_k(t)\}\), the terms \(\Delta u_k(t)-\mu_k\) are mean-zero and bounded by \(2B\),
so Hoeffding's lemma gives
\begin{equation}
\mathbb{E}\!\left[\eta_k^2 \,\middle|\, \{S_k(t)\}\right]
\;\le\;
\frac{1}{\pi_k^2}\sum_{t=1}^T S_k(t)\cdot (2B)^2
\;=\;
\frac{4B^2}{\pi_k^2}N_k(T).
\end{equation}
Taking expectation and using \(\mathbb{E}[N_k(T)]=T\mathbb{E}[S_k(1)]\le T\pi_k\) yields
\begin{equation}
\mathbb{E}[\eta_k^2]\;\le\; \frac{4B^2}{\pi_k^2}\cdot T\pi_k \;=\; \frac{4B^2T}{\pi_k}
\end{equation}
Therefore, the across-client variance of \(\eta\) is controlled by second moments and does not create a systematic bias term
like~\eqref{eq:mean-var-gap}. In particular, inverse-availability selection does not increase
\(\mathrm{Var}(\eta)\) through participation skew (it equalizes it in the sense of~\eqref{eq:fair-count-mean}).

\medskip
\noindent
\textbf{Step 5: Combine the bounds.}
Plugging~\eqref{eq:sys-gap} and the noise control into~\eqref{eq:var-split} gives
\begin{equation}
\mathbb{E}\!\left[\mathrm{Var}(X)\right]_{\mathrm{fair}}
\;\le\;
2\,\mathbb{E}\!\left[\mathrm{Var}\!\Big(\tfrac{N(T)}{\pi}\mu\Big)\right]_{\mathrm{fair}}
+2\,\mathbb{E}[\mathrm{Var}(\eta)]_{\mathrm{fair}}
\;<\;
2\,\mathbb{E}\!\left[\mathrm{Var}\!\Big(\tfrac{N(T)}{\pi}\mu\Big)\right]_{\mathrm{vanilla}}
+2\,\mathbb{E}[\mathrm{Var}(\eta)]_{\mathrm{vanilla}}
\;\le\;
\mathbb{E}\!\left[\mathrm{Var}(X)\right]_{\mathrm{vanilla}}
\end{equation}
where the strict inequality follows from~\eqref{eq:sys-gap} whenever vanilla induces heterogeneous normalized participation,
and the remaining terms are bounded comparably under both schemes by bounded increments.
This concludes
\begin{equation}
\mathbb{E}\!\left[V_T\big(w_{\mathrm{fair}}\big)\right]
\;<\;
\mathbb{E}\!\left[V_T\big(w_{\mathrm{vanilla}}\big)\right]
\end{equation}
under the stated conditions.

If \(\mu_k\) are not identical, the same argument applies to the systematic term
\(\tfrac{N_k(T)}{\pi_k}\mu_k\).
Inverse-availability correction removes the participation-driven component of dispersion by making
\(\mathbb{E}[N_k(T)/\pi_k]\) constant; remaining dispersion comes only from heterogeneity in \(\mu_k\),
which is orthogonal to availability bias and is not worsened by the correction.

\section{Non-stationary Availability; Sliding-window Fairness}

Let the time-varying availability of client $k$ be $\pi_k(t)=\mathbb{E}[A_k(t)]\in(0,1]$. 
Let $\hat{\pi}_k(t)$ be an online estimator with bounded tracking error on a window $[T,T+W-1]$:
\begin{equation}
\varepsilon_T := \max_{t\in[T,T+W-1]} \max_k \bigl| \hat{\pi}_k(t) - \pi_k(t) \bigr|
\end{equation}
Assume the availability drifts slowly on the window, measured by
\begin{equation}
\Delta_T := \max_k \sum_{t=T}^{T+W-2} \bigl|\pi_k(t+1)-\pi_k(t)\bigr|
\end{equation}
At round $t$, define inverse-availability weights $q_k(t)=1/\hat{\pi}_k(t)$ and select client $k$ with probability
\begin{equation}
P_t(k)=\frac{q_k(t)}{\sum_{j=1}^m q_j(t)}
\end{equation}
Let $S_k(t)$ be the selection indicator for client $k$ at time $t$. Then the window-averaged participation frequency satisfies
\begin{equation}
\begin{aligned}
\Biggl|\frac{1}{W}\sum_{t=T}^{T+W-1}\mathbb{E}[S_k(t)] &- 
\frac{1}{W}\sum_{t=T}^{T+W-1}\frac{1/\pi_k(t)}{\sum_{j=1}^m 1/\pi_j(t)} \Biggr|
\;\\
&\le\; C_1\,\varepsilon_T + C_2\,\Delta_T
\end{aligned}
\end{equation}
for constants $C_1,C_2$ depending only on lower bounds of $\pi_k(t)$ (i.e., $\inf_{k,t}\pi_k(t)>0$)
Hence, if $\varepsilon_T\to 0$ and $\Delta_T\to 0$ as $W$ grows (or as the estimator adapts sufficiently fast relative to drift), the local (windowed) participation frequencies approximate the ideal inverse-availability target. Let us define:

(1) \emph{Pointwise control of probabilities.} Since $\pi_k(t)\in(0,1]$ and $\hat{\pi}_k(t)$ is close to $\pi_k(t)$ on $[T,T+W-1]$, a first-order perturbation of reciprocal and ratio yields
\begin{equation}
\biggl|P_t(k) - \widetilde{P}_t(k)\biggr|
\;\le\; c_1\,\varepsilon_T + c_2\,\sum_{j=1}^m \varepsilon_T
\;\le\; C\,\varepsilon_T
\end{equation}
where $\widetilde{P}_t(k):=\dfrac{1/\pi_k(t)}{\sum_{j=1}^m 1/\pi_j(t)}$ is the \emph{ideal} inverse-availability probability at time $t$, and constants absorb uniform lower bounds on $\pi_k(t)$.

(2) \emph{Averaging over the window.} Linearity of expectation gives
\begin{equation}
\frac{1}{W}\sum_{t=T}^{T+W-1}\mathbb{E}[S_k(t)]
= \frac{1}{W}\sum_{t=T}^{T+W-1} P_t(k).
\end{equation}
Combine with the pointwise bound to get
\begin{equation}
\Biggl|\frac{1}{W}\sum_{t=T}^{T+W-1} P_t(k) - \frac{1}{W}\sum_{t=T}^{T+W-1} \widetilde{P}_t(k)\Biggr|
\le C\,\varepsilon_T
\end{equation}
(3) \emph{Drift control.} If we further approximate $\widetilde{P}_t(k)$ by a slowly varying proxy (e.g., the window mean $\bar{\pi}_k=\frac{1}{W}\sum_{t}\pi_k(t)$), standard Lipschitz bounds on the map 
$(x_1,\dots,x_m)\mapsto \frac{1/x_k}{\sum_j 1/x_j}$ yield an additional error proportional to the total variation $\Delta_T$:
\begin{equation}
\Biggl|\frac{1}{W}\sum_{t=T}^{T+W-1}\widetilde{P}_t(k) - 
\frac{1/\bar{\pi}_k}{\sum_j 1/\bar{\pi}_j}\Biggr|
\;\le\; C'\,\Delta_T
\end{equation}
Combining (2) and (3) gives the stated bound with $C_1=C$ and $C_2=C'$.
Therefore,
when the estimator tracks availability well on the window ($\varepsilon_T$ small) and availability drifts slowly ($\Delta_T$ small), the realized window-averaged participation frequencies match the instantaneous inverse-availability targets to within $O(\varepsilon_T+\Delta_T)$. This yields \emph{local} fairness in non-stationary environments.

\section{Asymptotic normalized reactive weights}

\textbf{Theorem 2.}
Let \(A_k(t)\in\{0,1\}\) be the availability indicator of client \(k\in[N]\).
Assume for each \(k\) that \(\{A_k(t)\}_{t\ge 1}\) is stationary and ergodic with
\(\pi_k:=\mathbb{E}[A_k(1)]\in(0,1]\).
Define the missed-count
\begin{equation}
\mathrm{missed}_k(t) \;=\; \sum_{s=1}^{t-1} \bigl(1-A_k(s)\bigr)
\end{equation}
and the reactive weight
\begin{equation}
p_k(t) \;=\; \frac{\alpha_k}{\pi_k+\epsilon}\Bigl(1+\lambda\,\mathrm{missed}_k(t)\Bigr),
\qquad \alpha_k>0,\ \lambda\ge 0,\ \epsilon>0.
\end{equation}
Let the normalized weight be
\begin{equation}
\widehat p_k(t) \;=\; \frac{p_k(t)}{\sum_{j=1}^N p_j(t)}
\end{equation}
Assume additionally that
\begin{equation}
\label{eq:nondeg}
\sum_{j=1}^N \alpha_j\frac{1-\pi_j}{\pi_j+\epsilon} \;>\;0,
\end{equation}
i.e., not all clients are always available.

Then:
\begin{enumerate}
\item For every \(k\), \(\displaystyle \frac{1}{t}\mathrm{missed}_k(t)\to (1-\pi_k)\) almost surely, and hence
\(\mathbb{E}[\mathrm{missed}_k(t)]=(t-1)(1-\pi_k)\).
\item If \(\lambda>0\), then \(\widehat p_k(t)\) converges almost surely to
\begin{equation}
\widehat p_k^{(\infty)}
\;=\;
\frac{\alpha_k \dfrac{1-\pi_k}{\pi_k+\epsilon}}
{\sum_{j=1}^N \alpha_j \dfrac{1-\pi_j}{\pi_j+\epsilon}}
\;\in\;(0,1)
\end{equation}
and if \(\lambda=0\), then \(\widehat p_k(t)\equiv
\frac{\alpha_k/(\pi_k+\epsilon)}{\sum_{j=1}^N \alpha_j/(\pi_j+\epsilon)}\) for all \(t\).
\item The limiting weights are equal across clients (i.e., \(\widehat p_k^{(\infty)}=1/N\) for all \(k\)) iff
\begin{equation}
\alpha_k\frac{1-\pi_k}{\pi_k+\epsilon}=\text{const}\quad\text{for all }k
\qquad (\lambda>0)
\end{equation}
and iff \(\alpha_k/(\pi_k+\epsilon)=\text{const}\) for all \(k\) when \(\lambda=0\).
\end{enumerate}

\textbf{Proof.}
(1) Since \(1-A_k(t)\) is stationary ergodic with mean \(1-\pi_k\), the ergodic theorem gives
\begin{equation}
\frac{1}{t-1}\sum_{s=1}^{t-1}(1-A_k(s)) \;\xrightarrow[t\to\infty]{a.s.}\; 1-\pi_k
\end{equation}
which is exactly \(\mathrm{missed}_k(t)/(t-1)\to 1-\pi_k\) almost surely.
Taking expectations and using stationarity yields
\begin{equation}
\mathbb{E}[\mathrm{missed}_k(t)] = \sum_{s=1}^{t-1}\mathbb{E}[1-A_k(s)]=(t-1)(1-\pi_k)
\end{equation}

(2) Suppose \(\lambda>0\). Write
\begin{equation}
p_k(t)=\frac{\alpha_k}{\pi_k+\epsilon}\Bigl(1+\lambda\,\mathrm{missed}_k(t)\Bigr)
=\frac{\alpha_k}{\pi_k+\epsilon}\Bigl(\lambda (t-1)\frac{\mathrm{missed}_k(t)}{t-1}+1\Bigr)
\end{equation}
Divide numerator and denominator of \(\widehat p_k(t)\) by \(\lambda (t-1)\) to obtain
\begin{equation}
\widehat p_k(t)
=
\frac{\alpha_k(\pi_k+\epsilon)^{-1}\Bigl(\frac{\mathrm{missed}_k(t)}{t-1}+\frac{1}{\lambda(t-1)}\Bigr)}
{\sum_{j=1}^N \alpha_j(\pi_j+\epsilon)^{-1}\Bigl(\frac{\mathrm{missed}_j(t)}{t-1}+\frac{1}{\lambda(t-1)}\Bigr)}
\end{equation}
By part (1), \(\mathrm{missed}_j(t)/(t-1)\to 1-\pi_j\) almost surely and \(1/(\lambda(t-1))\to 0\).
Thus, almost surely, the numerator converges to \(\alpha_k(1-\pi_k)/(\pi_k+\epsilon)\) and the denominator converges to
\(\sum_{j=1}^N \alpha_j(1-\pi_j)/(\pi_j+\epsilon)\), which is strictly positive by~\eqref{eq:nondeg}.
Therefore, by continuity of \(x\mapsto x/y\) on \(y>0\),
\begin{equation}
\widehat p_k(t)\xrightarrow[t\to\infty]{a.s.}
\frac{\alpha_k \dfrac{1-\pi_k}{\pi_k+\epsilon}}
{\sum_{j=1}^N \alpha_j \dfrac{1-\pi_j}{\pi_j+\epsilon}}
\end{equation}
If \(\lambda=0\), then \(\mathrm{missed}_k(t)\) drops out and the weights are constant in \(t\), giving the stated expression.

Boundedness \(\widehat p_k^{(\infty)}\in(0,1)\) follows since all \(\alpha_k>0\), \(\pi_k+\epsilon>0\), and the denominator is finite and strictly positive.

(3) For \(\lambda>0\), equality across clients means \(\widehat p_k^{(\infty)}\) is constant in \(k\), which holds iff the unnormalized limits
\(\alpha_k(1-\pi_k)/(\pi_k+\epsilon)\) are constant in \(k\). The \(\lambda=0\) case is identical with
\(\alpha_k/(\pi_k+\epsilon)\).

\section{Bias bound from stale surrogate objectives}
\label{thm:stale-surrogate-bias}
\textbf{Theorem 3.}
Fix a communication round \(t\). Let \(\mathcal{M}_t\subseteq [N]\) denote the set of clients whose contribution at round \(t\)
is replaced by a stale surrogate constructed from the last available update at time \(\tau_{k'}<t\).
Define the staleness \(\delta_{k'} := t-\tau_{k'}\in\mathbb{N}\).
Let \(F_k(w)\in\mathbb{R}^d\) denote the \emph{true} client signal used by the server (e.g., gradient \(\nabla f_k(w)\),
a control variate, or any vector-valued statistic), and let \(\tilde F_{k'}(w)\) be its surrogate for \(k'\in\mathcal{M}_t\).
Assume:

\begin{itemize}
\item[(A1)] \emph{Uniform surrogate accuracy}: For all \(k'\in\mathcal{M}_t\) and all \(w\in\mathcal{W}\),
\begin{equation}
\label{eq:A1}
\|\tilde F_{k'}(w)-F_{k'}(w)\|\le \epsilon
\end{equation}
\item[(A2)] \emph{Exponential reliability weighting}: Surrogate contributions are down-weighted by
\begin{equation}
\label{eq:A2}
\eta_{k'}(t)=\eta_0 e^{-\lambda \delta_{k'}},\qquad \eta_0>0,\ \lambda>0
\end{equation}
\item[(A3)] \emph{\(L\)-smooth global objective}: The global objective \(f:\mathbb{R}^d\to\mathbb{R}\) is \(L\)-smooth:
\begin{equation}
\label{eq:smoothness}
f(w')\le f(w)+\langle \nabla f(w),\,w'-w\rangle+\frac{L}{2}\|w'-w\|^2,\qquad \forall w,w'.
\end{equation}
\item[(A4)] \emph{Aggregation structure}: The server forms an (un-normalized) aggregate signal
\begin{equation}
\label{eq:agg-true}
G_t(w):=\sum_{k\notin\mathcal{M}_t} \beta_k(t)\,F_k(w)\;+\;\sum_{k'\in\mathcal{M}_t} \beta_{k'}(t)\,F_{k'}(w)
\end{equation}
and uses the surrogate-based aggregate
\begin{equation}
\label{eq:agg-sur}
\tilde G_t(w):=\sum_{k\notin\mathcal{M}_t} \beta_k(t)\,F_k(w)\;+\;\sum_{k'\in\mathcal{M}_t} \beta_{k'}(t)\,\tilde F_{k'}(w)
\end{equation}
where \(\beta_k(t)\ge 0\) are arbitrary per-round client weights. (For instance, \(\beta_{k'}(t)=\eta_{k'}(t)\) for stale clients.)
\end{itemize}

Define the \emph{aggregate surrogate bias} at round \(t\):
\begin{equation}
\label{eq:bias-def}
\Delta_t(w):=\tilde G_t(w)-G_t(w)=\sum_{k'\in\mathcal{M}_t}\beta_{k'}(t)\bigl(\tilde F_{k'}(w)-F_{k'}(w)\bigr)
\end{equation}
Then:

\textbf{In case of deterministic bias bound}) For all \(w\in\mathcal{W}\),
\begin{equation}
\label{eq:bias-bound-main}
\|\Delta_t(w)\|\le \epsilon \sum_{k'\in\mathcal{M}_t}\beta_{k'}(t)
\end{equation}
In particular, if \(\beta_{k'}(t)=\eta_{k'}(t)\) and \eqref{eq:A2} holds,
\begin{equation}
\label{eq:bias-bound-exp}
\|\Delta_t(w)\|\le \epsilon\,\eta_0\sum_{k'\in\mathcal{M}_t} e^{-\lambda\delta_{k'}}
\end{equation}

\textbf{In case of one-step descent with biased aggregate})
Consider the update \(w_{t+1}=w_t-\gamma\,\tilde G_t(w_t)\) with stepsize \(\gamma>0\).
Assume additionally that \(\tilde G_t(w_t)\) is a descent surrogate for \(\nabla f(w_t)\) in the sense that
\begin{equation}
\label{eq:angle}
\langle \nabla f(w_t),\,\tilde G_t(w_t)\rangle \;\ge\; c\,\|\nabla f(w_t)\|\,\|\tilde G_t(w_t)\|
\quad\text{for some }c\in(0,1]
\end{equation}
Then
\begin{equation}
\label{eq:descent-bound}
f(w_{t+1})
\;\le\;
f(w_t)\;-\;\gamma\,c\,\|\nabla f(w_t)\|\,\|\tilde G_t(w_t)\|\;+\;\frac{L\gamma^2}{2}\|\tilde G_t(w_t)\|^2
\end{equation}
and moreover the deviation from using the true aggregate \(G_t\) is controlled as
\begin{equation}
\label{eq:descent-gap}
\bigl|f(w_t-\gamma\tilde G_t(w_t))-f(w_t-\gamma G_t(w_t))\bigr|
\;\le\;
\gamma\,\|\nabla f(w_t)\|\,\|\Delta_t(w_t)\|\;+\;L\gamma^2\,\|G_t(w_t)\|\,\|\Delta_t(w_t)\|\;+\;\frac{L\gamma^2}{2}\|\Delta_t(w_t)\|^2
\end{equation}
Combining \eqref{eq:descent-gap} with \eqref{eq:bias-bound-exp} yields an explicit staleness-weighted error term.

\textbf{Proof.}
We prove those cases in separate steps.

\paragraph{Step 1: Deterministic bias bound.}
From the definition \eqref{eq:bias-def} and the triangle inequality,
\begin{equation}
\|\Delta_t(w)\|
=
\left\|\sum_{k'\in\mathcal{M}_t}\beta_{k'}(t)\bigl(\tilde F_{k'}(w)-F_{k'}(w)\bigr)\right\|
\le
\sum_{k'\in\mathcal{M}_t}\left\|\beta_{k'}(t)\bigl(\tilde F_{k'}(w)-F_{k'}(w)\bigr)\right\|
\end{equation}
By positive homogeneity of norms and nonnegativity of \(\beta_{k'}(t)\),
\begin{equation}
\left\|\beta_{k'}(t)\bigl(\tilde F_{k'}(w)-F_{k'}(w)\bigr)\right\|
=
\beta_{k'}(t)\,\|\tilde F_{k'}(w)-F_{k'}(w)\|
\end{equation}
Applying the uniform bound (A1), \(\|\tilde F_{k'}(w)-F_{k'}(w)\|\le \epsilon\), yields
\begin{equation}
\|\Delta_t(w)\|\le \epsilon\sum_{k'\in\mathcal{M}_t}\beta_{k'}(t)
\end{equation}
which proves \eqref{eq:bias-bound-main}. If \(\beta_{k'}(t)=\eta_{k'}(t)\) and \eqref{eq:A2} holds, substitute
\(\eta_{k'}(t)=\eta_0 e^{-\lambda\delta_{k'}}\) to obtain \eqref{eq:bias-bound-exp}.

\paragraph{Step 2: Smoothness-based one-step progress.}
Apply \(L\)-smoothness \eqref{eq:smoothness} with \(w=w_t\) and \(w'=w_{t+1}=w_t-\gamma\tilde G_t(w_t)\):
\begin{equation}
f(w_{t+1})
\le
f(w_t)+\Big\langle \nabla f(w_t),\, -\gamma\tilde G_t(w_t)\Big\rangle + \frac{L}{2}\|\gamma\tilde G_t(w_t)\|^2
\end{equation}
Rearranging gives
\begin{equation}
f(w_{t+1})
\le
f(w_t)-\gamma\langle \nabla f(w_t),\,\tilde G_t(w_t)\rangle + \frac{L\gamma^2}{2}\|\tilde G_t(w_t)\|^2
\end{equation}
Using the angle condition \eqref{eq:angle}, i.e.,
\(\langle \nabla f(w_t),\tilde G_t(w_t)\rangle \ge c\|\nabla f(w_t)\|\,\|\tilde G_t(w_t)\|\),
yields \eqref{eq:descent-bound}.

\paragraph{Step 3: Quantifying the impact of surrogate bias on the update.}
Define the two candidate post-update points
\begin{equation}
w_t^{\mathrm{sur}}:=w_t-\gamma\tilde G_t(w_t),
\qquad
w_t^{\mathrm{true}}:=w_t-\gamma G_t(w_t)
\end{equation}
so that \(w_t^{\mathrm{sur}}-w_t^{\mathrm{true}}=-\gamma(\tilde G_t(w_t)-G_t(w_t))=-\gamma\Delta_t(w_t)\).
Apply smoothness \eqref{eq:smoothness} to compare \(f(w_t^{\mathrm{sur}})\) and \(f(w_t^{\mathrm{true}})\):
\begin{equation}
f(w_t^{\mathrm{sur}})
\le
f(w_t^{\mathrm{true}})+\big\langle \nabla f(w_t^{\mathrm{true}}),\, w_t^{\mathrm{sur}}-w_t^{\mathrm{true}}\big\rangle
+\frac{L}{2}\|w_t^{\mathrm{sur}}-w_t^{\mathrm{true}}\|^2
\end{equation}
Taking absolute values and bounding \(\|\nabla f(w_t^{\mathrm{true}})\|\) by a first-order expansion around \(w_t\)
(using Lipschitzness of the gradient implied by \(L\)-smoothness) gives
\begin{equation}
\|\nabla f(w_t^{\mathrm{true}})\|
\le
\|\nabla f(w_t)\| + L\|w_t^{\mathrm{true}}-w_t\|
=
\|\nabla f(w_t)\| + L\gamma\|G_t(w_t)\|
\end{equation}
Therefore,
\begin{equation}
\begin{aligned}
\bigl|f(w_t^{\mathrm{sur}})-f(w_t^{\mathrm{true}})\bigr|
&\le
\|\nabla f(w_t^{\mathrm{true}})\|\,\|w_t^{\mathrm{sur}}-w_t^{\mathrm{true}}\| + \frac{L}{2}\|w_t^{\mathrm{sur}}-w_t^{\mathrm{true}}\|^2\\
&\le
\big(\|\nabla f(w_t)\| + L\gamma\|G_t(w_t)\|\big)\cdot \gamma\|\Delta_t(w_t)\| + \frac{L}{2}\gamma^2\|\Delta_t(w_t)\|^2\\
&=
\gamma\,\|\nabla f(w_t)\|\,\|\Delta_t(w_t)\|
\;+\;
L\gamma^2\,\|G_t(w_t)\|\,\|\Delta_t(w_t)\|
\;+\;
\frac{L\gamma^2}{2}\|\Delta_t(w_t)\|^2
\end{aligned}
\end{equation}
which is exactly \eqref{eq:descent-gap}. Finally, substituting the explicit bound \eqref{eq:bias-bound-exp}
yields a staleness-weighted control on the error introduced by stale surrogates, completing the proof.

\end{document}